\documentclass{article}

\PassOptionsToPackage{numbers, compress}{natbib}
\usepackage[main, final]{neurips_2026}

\usepackage[utf8]{inputenc} 
\usepackage[T1]{fontenc}    
\usepackage{hyperref}       
\usepackage{url}            
\usepackage{booktabs}       
\usepackage{amsfonts}       
\usepackage{nicefrac}       
\usepackage{microtype}      
\usepackage{xcolor}         
\usepackage{graphicx}
\usepackage{amsmath}
\usepackage{subcaption}
\usepackage{caption}
\definecolor{best_color}{rgb}{1, 0.7, 0.7}
\definecolor{second_color}{rgb}{1, 0.85, 0.7}
\definecolor{third_color}{rgb}{1, 1, 0.7}

\title{DirectUV: Image-Conditioned UV Texture Generation with Surface-Aware Positional Encoding}

\author{%
  Jiantao Lin$^{1,*}$ \quad
  Yingjie Xu$^{1,3*}$ \quad
  Mingzhi Sheng$^{1}$ \\
  \textbf{Yangkai Wei$^{3}$ \quad
  Hao Chen$^{2}$ \quad
  Ying-Cong Chen$^{1,2,3,\dagger}$} \\[0.5em]
  $^{1}$The Hong Kong University of Science and Technology (Guangzhou) \\
  $^{2}$The Hong Kong University of Science and Technology \\
  $^{3}$knowin.ai \\[0.25em]
  $^{*}$Equal contribution. \quad $^{\dagger}$Corresponding author.
}

\begin{document}

\maketitle

\begin{abstract}
Generating high-quality UV textures for 3D meshes remains challenging. Multi-view projection pipelines suffer from occlusion and view inconsistency, and recent methods that generate textures directly in UV space still rely on auxiliary modules to supply 3D information, leaving the attention mechanism tied to UV-grid positions rather than to the underlying surface geometry. This mismatch limits coherence across seams and disconnected UV islands.
We propose DirectUV, an image-conditioned UV texture diffusion framework that operates in the latent UV space of a pretrained image VAE, in which a Diffusion Transformer denoises the UV latent given a single input image and a coarse UV map. At its core, Surface-Aware Positional Encoding (SAPE) replaces the standard 2D-grid positional encoding with encodings derived from per-token 3D surface coordinates obtained via UV-to-surface correspondence. As positional encodings define the distance metric that attention operates on, SAPE enables tokens to attend to each other based on true surface proximity rather than UV-grid distance, restoring coherence across seams and disconnected islands. A multi-level extension further assigns different attention heads to progressively finer subdivisions of the same latent UV patch, allowing the model to reason about surface structure at multiple granularities.
Experiments show that DirectUV produces sharper and more globally consistent textures than other baselines, with the largest improvements in occluded and view-unseen regions where projection-based methods leave gaps or stretched textures.
\end{abstract}

\section{Introduction}
Generating high-quality textures for 3D meshes requires reconciling two objectives: producing realistic and detailed appearance, and ensuring consistency across the entire 3D surface, free from seams, view-dependent artifacts, and misaligned details.

Existing approaches to generative mesh texturing largely fall into two paradigms. The first leverages powerful 2D diffusion priors to synthesize multi-view images, which are then projected and baked onto the mesh surface~\cite{chen2023fantasia3d,richardson2023texture,chen2023text2tex,zeng2024paint3d,cheng2025mvpaint,yan2024flexipainter,feng2025romantex,georgiou2025im2surftex}. While effective at producing realistic local appearance, these methods suffer from multi-view inconsistency and error accumulation during projection, often resulting in seams, misalignment, and unstable textures. The second paradigm generates textures directly in UV space or mesh-native representations~\cite{yu2023texture,yu2024texgen,liang2025UnitTEX}, avoiding projection artifacts by construction. While some of these methods incorporate 3D information through auxiliary modules such as point cloud attention layers~\cite{yu2024texgen} or cross-attention with surface coordinates~\cite{liu2025texgarment}, the core attention mechanism still operates on UV-space positions, leaving the model's notion of token proximity tied to the 2D UV layout rather than the underlying 3D surface geometry. As a result, texels that are adjacent on the surface but separated across UV charts remain difficult to relate, limiting coherence around seams and across islands. Recent attempts to extend positional encoding itself with 3D awareness, including 3D-aware rotary embeddings for view-space generation~\cite{feng2025romantex,hunyuan3d2025hunyuan3d} and hierarchical multi-level encodings for general 3D fields~\cite{bai2025positional}, are either tied to the multi-view image grid or designed for tasks beyond UV texture synthesis, leaving open the question of how to install surface geometry into UV-space attention itself.

Unlike conditioning signals that the model may selectively attend to, positional encodings define the distance metric that attention is compelled to operate on, making them the natural place to embed surface geometry. Recent analysis of diffusion transformers shows that spatial coherence among patch tokens is primarily governed by positional encodings rather than by token-to-token content interactions~\cite{bai2025positional}, reinforcing positional encoding as the structural locus where spatial priors take effect.
This observation suggests that, rather than relying on auxiliary modules to supply 3D information, surface geometry should be embedded directly into the attention mechanism. We introduce DirectUV, a diffusion-based framework that generates textures directly in latent UV space, conditioned on a single input image and a coarse UV map, while encoding 3D surface geometry into the positional encodings of the diffusion transformer.

DirectUV operates in the latent space of a pretrained image VAE, with the UV map as the sole generation target. A single input image provides appearance guidance, and a coarse UV map supplies surface-aligned layout cues. Both serve purely as conditioning signals rather than intermediate textures to be refined, preventing view-dependent artifacts from propagating into the final output.
To bridge the gap between the 2D UV layout and the underlying 3D surface, we introduce Surface-Aware Positional Encoding (SAPE). Unlike existing methods that supply 3D information through auxiliary conditioning modules, SAPE embeds per-token 3D surface coordinates, obtained through UV-to-surface correspondence, directly into the positional encodings of the diffusion transformer. By redefining the positional encoding in this way, tokens attend to each other based on relative offsets between their ambient 3D coordinates, rather than receiving 3D geometry as an external signal that the model may or may not learn to follow. As a result, surface-adjacent texels can be directly related in attention even when they are disconnected in the UV layout, without requiring additional architectural components.
We further extend SAPE to \textbf{Multi-Level SAPE}, which exploits the multi-head structure of the transformer by assigning different attention heads positional encodings derived from progressively finer subdivisions of the same latent UV patch. As a result, the model can represent both patch-level surface correspondence and finer sub-patch geometric variation within the existing multi-head attention framework, without introducing additional parameters or modules.

Our contributions are as follows:
\begin{itemize}
\item We formulate UV texture generation as a direct image-conditioned generation problem in latent UV space, where a single input image and a coarse UV map serve purely as conditioning signals, rather than generation targets to be projected or refined.
\item We introduce Surface-Aware Positional Encoding (SAPE), which embeds per-token 3D surface coordinates into the positional encodings of the diffusion transformer, enabling attention to operate on relative offsets between ambient 3D coordinates rather than UV-grid distance. We further introduce Multi-Level SAPE, which leverages the multi-head structure through progressively finer subdivisions of the same latent UV patch, allowing different heads to capture surface geometry at multiple positional granularities.
\item Experiments demonstrate that DirectUV produces sharper, more seam-consistent textures than both projection-based and direct-generation baselines, with the largest gains in occluded and view-unseen regions that are poorly covered by camera-based projection.
\end{itemize}

\section{Related Work}

\subsection{Multi-View Projection-Based Texturing} 
A dominant line of work leverages the strong image prior of large pretrained 2D diffusion models~\cite{flux2024,su2024roformer} to synthesize textured renderings from multiple viewpoints, typically conditioned on geometry cues such as depth or normals, and then consolidates them into a surface texture via projection, baking, and inpainting. Representative pipelines include~\cite{liu2023zero1to3,huang2025mv,long2024wonder3d,lin2025kiss3dgen,yang2025advancing,liu2023text}. This paradigm is appealing as it reuses powerful 2D diffusion backbones with limited 3D supervision, while naturally supporting text prompts and reference images.

However, inconsistencies in texture detail and illumination remain difficult to fully eliminate, even with joint multi-view denoising. A finite set of viewpoints cannot cover occluded or grazing-angle regions, and the subsequent projection and baking must reconcile mismatched signals across views, leading to accumulated errors such as seam artifacts and local collapse. Recent work has begun incorporating 3D awareness directly into the attention of view-space diffusion models, such as the Paint module in Hunyuan3D-2.1~\cite{hunyuan3d2025hunyuan3d}, which adopts the 3D-aware RoPE from RomanTex~\cite{feng2025romantex}. Nevertheless, these methods still rely on a projection-and-baking stage to map synthesized views onto the mesh surface.

\subsection{Direct Texture Generation in UV Space}
An alternative line of work avoids multi-view projection by generating textures directly in UV space or other mesh-native representations, eliminating projection artifacts by construction. Point-UV Diffusion~\cite{yu2023texture} adopts a coarse-to-fine pipeline combining point cloud representations with UV-space diffusion to improve structural consistency. TEXGen~\cite{yu2024texgen} scales this direction with a large diffusion model that interleaves UV-space convolutions and point cloud attention, enabling feed-forward generation of high-resolution textures conditioned on single-view image and text. TexGarment~\cite{liu2025texgarment} further incorporates global 3D structure via cross-attention between UV latents and point cloud features. More recently, UniTEX~\cite{liang2025UnitTEX} retains a multi-view projection stage but performs completion in a learned 3D feature space using a triplane  cube representation, querying unobserved regions and merging them with projected partial textures, this avoids UV chart discontinuities but treats projected pixels as ground truth to preserve rather than as conditioning to be regenerated. 

Across these methods, 3D structure is introduced through auxiliary mechanisms rather than being integrated into UV-space modeling itself. Point-UV Diffusion, TEXGen, and TexGarment inject geometry via point cloud features or cross-attention, leaving the UV diffusion process unaware of mesh topology at its core, while UniTEX shifts completion into a 3D feature space instead of enriching UV-space modeling. In contrast, DirectUV remains entirely in UV space and embeds surface geometry directly into the positional encodings of the diffusion transformer, making the diffusion process itself surface-aware without relying on auxiliary modules or non-UV representations.

\subsection{Positional Encoding in Diffusion Transformers}


Diffusion Transformers (DiTs)~\cite{peebles2023scalable} have emerged as a scalable backbone for high-quality image generation by operating on latent patch tokens. Within this architecture, positional encoding defines the spatial organization of tokens by shaping how attention relates different regions in latent space. Rotary positional encoding (RoPE)~\cite{su2024roformer}, widely adopted in modern generative models, injects positional information directly into query--key interactions and therefore determines the distance metric on which attention operates. Unlike conditioning signals that may be selectively attended to or ignored, positional encoding fundamentally governs token-to-token interactions throughout the denoising process, making it a natural place to encode spatial priors. Recent study~\cite{bai2025positional} have further shown that the design of positional encoding significantly affects spatial coherence, controllability, and generalization behavior in diffusion transformers, rather than serving merely as an implementation detail.

Recent work has begun extending positional encoding with explicit 3D awareness. RomanTex~\cite{feng2025romantex} introduces 3D-aware rotary embeddings for multi-view image diffusion, and Hunyuan3D-2.1~\cite{hunyuan3d2025hunyuan3d} adopts this design in its Paint module to improve consistency during multi-view texture synthesis. However, these methods still operate on multi-view image grids, where surface-aware information is consumed during view-space denoising and only later projected back onto the mesh surface. As a result, coherence across UV seams and disconnected surface regions must still be recovered during the projection and baking stage. In contrast, DirectUV operates directly on UV-space latent tokens, allowing surface-aware attention to participate in the texture generation process itself. This enables seam-crossing surface relationships to be resolved during denoising rather than deferred to a subsequent projection step.


\section{Method}\label{sec:method}
\subsection{Overview}
\begin{figure}[t]
\centering
\includegraphics[width=\linewidth]{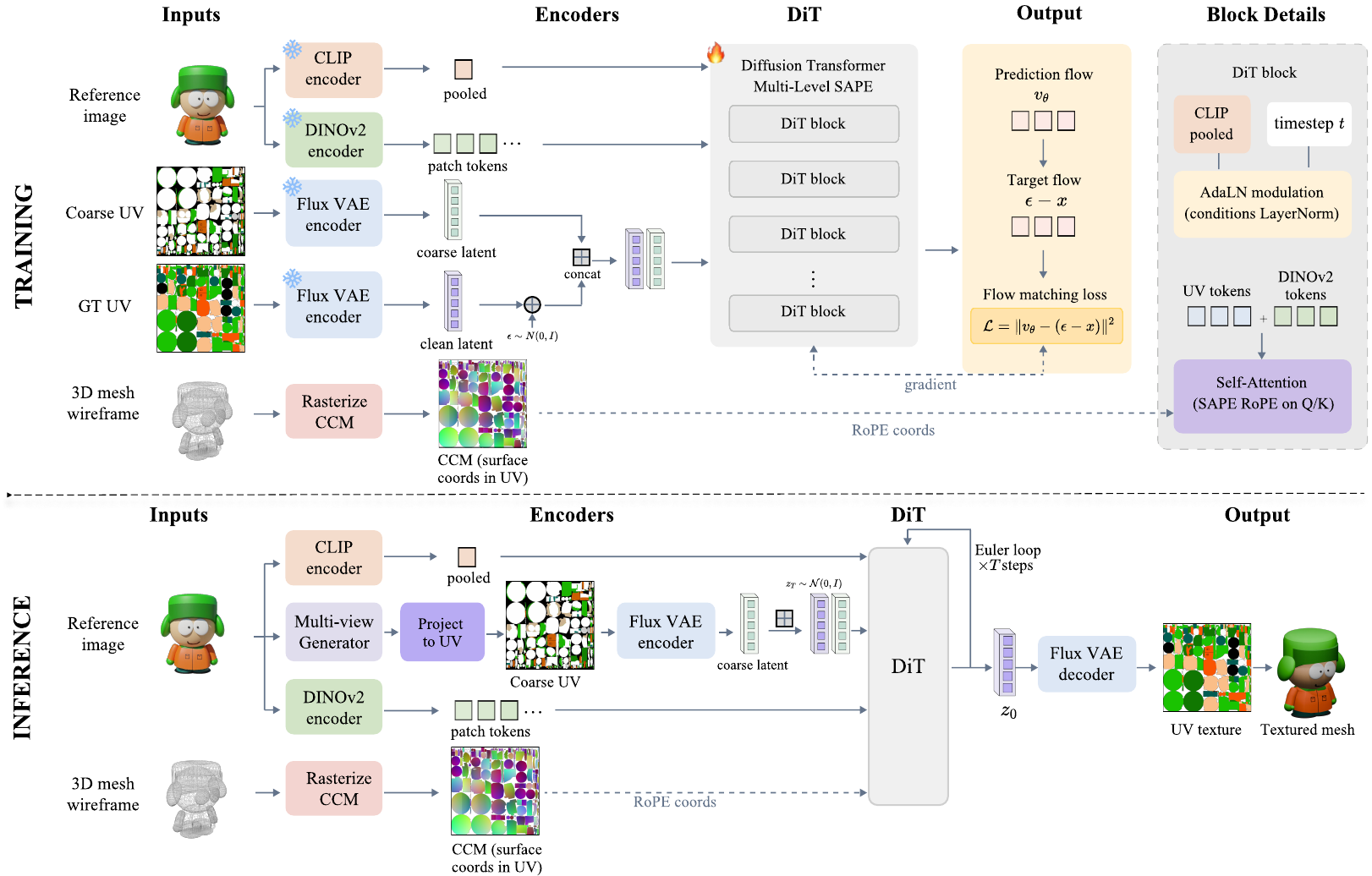}
\caption{Overview of DirectUV. \textbf{Top (training).} The reference image is encoded by CLIP~\cite{Radford2021LearningTV} into a pooled feature that drives AdaLN modulation, and by DINOv2~\cite{oquab2023dinov2} into patch tokens that feed the joint-attention stream. A coarse UV map is encoded by the frozen Flux VAE and channel-concatenated with the noisy UV latent to form the DiT input. The CCM supplies per-head RoPE coordinates for Multi-Level SAPE. The model is trained with the rectified flow-matching objective. \textbf{Bottom (inference).} The reference image drives both an off-the-shelf multi-view generator, whose outputs are projected to a coarse UV map, and the CLIP/DINOv2 encoders. Under this conditioning, $N$ Euler steps denoise $\mathbf{z}_1$ to $\mathbf{z}_0$, which is decoded by the frozen Flux VAE and wrapped onto the mesh as the final texture.}
\vspace{-1.8em}
\label{fig:pipeline}
\end{figure}

Since positional encoding governs the distance metric of attention, embedding surface coordinates into PE offers a direct path to surface-aware denoising without auxiliary 3D modules. DirectUV formulates image-conditioned UV texture generation as a diffusion process directly in UV space, with the UV map as the sole generation target. The pipeline operates in a latent UV space. Given an input image and a mesh with UV parameterization, the UV map is encoded into a compact latent representation by a pretrained image VAE. The same UV parameterization additionally yields a Canonical Coordinate Map (CCM), a UV-space image whose value at each pixel is the 3D surface coordinate of the corresponding mesh point. The CCM is precomputed once per mesh and provides the geometric input from which Surface-Aware Positional Encoding (SAPE) derives per-token surface coordinates. A diffusion transformer (DiT) then models the conditional generation process in this latent space, guided by the input image and a coarse UV map, with SAPE integrated into its attention layers to inject 3D surface geometry directly into token-to-token interactions. The predicted latent is finally decoded back to obtain the UV texture. An overview of the full training and inference pipeline is shown in Figure~\ref{fig:pipeline}.
In the following, we first introduce the latent UV diffusion formulation, then describe the conditioning strategy for incorporating image and UV cues, and finally present the design of SAPE for enabling surface-aware attention in UV-space generation.

\subsection{Latent UV Diffusion Formulation}
We model UV texture generation as a conditional diffusion process in a latent UV space. A UV map, defined as an image-like signal over a regular 2D grid, is encoded into a spatially downsampled latent representation $\mathbf{x} = \mathcal{E}(I_{\text{UV}})$ using a frozen Flux~\cite{flux2024} VAE encoder, on which all denoising is performed, and decoded back by $\mathcal{D}$ to obtain the final texture. This representation naturally aligns with pretrained 2D image priors, while the VAE reconstructs UV maps with negligible error, allowing the model to operate in a compact space without sacrificing fine-grained appearance.

With the UV map as the sole generation target, texture synthesis reduces to a single diffusion process, avoiding the inconsistencies introduced by projecting and blending intermediate views. The latent representation preserves the UV coordinate structure, providing a direct path for surface geometry to enter through positional encoding rather than through auxiliary modules or external 3D features. We instantiate $v_\theta$ using the Flux DiT architecture, and introduce the conditioning bundle $\mathbf{c}$ together with Surface-Aware Positional Encoding (SAPE) within its attention layers, whose details are presented in the following subsections.

\subsection{Conditioning in UV-Space Generation}\label{sec:cond}

The conditioning bundle $\mathbf{c}$ consists of two complementary signals: a reference image providing appearance information and a coarse UV map encoding surface-aligned layout. These signals are injected into the DiT through different pathways depending on whether they share the UV coordinate system of the target texture.

\noindent\textbf{Image condition.}  
The reference image is incorporated through two streams aligned with how the DiT processes external inputs. A pooled CLIP feature captures global appearance attributes such as color, material tone, and identity, and is injected through the modulation pathway used for timestep and text embeddings in Flux, conditioning all tokens uniformly. In parallel, dense patch tokens from DINOv2 provide spatially localized details and are fed into the joint-attention stream, where they effectively replace text tokens in the original architecture. This enables UV latent tokens to attend to relevant image regions for fine-grained appearance cues.

\noindent\textbf{Coarse UV map.}  
The coarse UV map is obtained by projecting one or more input views into UV space and provides surface-aligned layout guidance. Since it shares the UV coordinate system with the target texture, it is injected through a position-aligned pathway. The same Flux VAE encodes it into a latent representation, which is concatenated with the noisy UV latent along the channel dimension before patchification. This per-position concatenation preserves one-to-one spatial correspondence with the latent being denoised, unlike the image condition, which relies on modulation and cross-attention due to the lack of coordinate alignment.

In contrast to projection-based pipelines that treat projected UV maps as intermediate results for refinement or inpainting, both signals act purely as conditioning. The UV texture is generated from scratch in latent space, allowing DirectUV to leverage multi-view cues without inheriting view-dependent inconsistencies.

\subsection{Surface-Aware Positional Encoding (SAPE)}
\noindent\textbf{Spatial Metric Mismatch.}
In a diffusion transformer, positional encodings define the spatial metric over which attention relates tokens, fixing which regions are treated as neighbors during denoising. For UV texture generation this metric is critical: appearance should propagate according to proximity on the mesh surface rather than proximity on the flattened UV plane.

Standard DiTs use positional encodings on a regular 2D grid, so token relationships are measured in UV coordinates. This induces a mismatch between the parameterization and the underlying surface: texels that are immediate neighbors on the mesh can be split across a UV seam onto disconnected islands and end up arbitrarily far apart under the UV-grid metric (Figure~\ref{fig:sape_mechanism}, left), while texels nearby in the UV plane may correspond to distant surface regions. The 2D grid therefore supplies attention with an incorrect neighborhood prior, limiting coherence across seams and surface-disconnected UV regions.

\noindent\textbf{Surface-Aware RoPE.}
We address this mismatch with \textbf{Surface-Aware Positional Encoding (SAPE)}, which replaces the 2D UV-grid positional encoding with positional codes derived from the CCM. Let $\mathbf{x}_i$ denote the $i$-th token of the latent $\mathbf{x}$ and $U_i$ the UV region it covers. We downsample the CCM to the resolution of the token grid so that $U_i$ maps to a single CCM pixel, whose value $\mathcal{S}_i \in \mathbb{R}^3$ gives the per-token surface coordinate. For any pair of tokens $\mathbf{x}_i, \mathbf{x}_j$ at head $h$, SAPE rotates their query and key as
\begin{equation}
\tilde{\mathbf{Q}}_i^{(h)} = \mathrm{RoPE}(\mathbf{Q}_i^{(h)}; \mathcal{S}_i), \qquad
\tilde{\mathbf{K}}_j^{(h)} = \mathrm{RoPE}(\mathbf{K}_j^{(h)}; \mathcal{S}_j).
\end{equation}
Following the standard axis-factorized form of RoPE, the head's feature dimension is split into three contiguous groups corresponding to the $(x, y, z)$ axes of $\mathcal{S}_i$, with each group rotated independently by its own coordinate. The dot product $\langle \tilde{\mathbf{Q}}_i^{(h)}, \tilde{\mathbf{K}}_j^{(h)} \rangle$ therefore depends only on the relative offset $\mathcal{S}_j - \mathcal{S}_i$ on the 3D surface rather than on the 2D UV grid. Texels nearby on the surface can therefore attend to each other directly, even when separated in UV space.

\begin{minipage}{0.48\textwidth}
\noindent\textbf{Multi-Level SAPE.}
A single per-token surface coordinate cannot simultaneously capture broad surface correspondence across UV islands and fine local structure around seams. We therefore extend SAPE to a multi-level form that reads the CCM at $L$ resolutions over the same latent patch. At level $l \in \{0, \ldots, L-1\}$, the patch is subdivided into a $2^l \times 2^l$ layout, yielding $4^l$ sub-cell surface coordinates per token. We denote any one of them by $\mathcal{S}_i^{(l)}$, so the basic SAPE above corresponds to $l=0$ with $\mathcal{S}_i = \mathcal{S}_i^{(0)}$.

These per-level coordinates are then distributed across the head dimension of the diffusion transformer. Multi-head attention partitions the per-token feature into $H$ contiguous slices of size $d_\text{head}$, with each head rotating its own slice independently. Each head $h$ is assigned a level $l_h \in \{0, \ldots, L-1\}$ and a specific sub-cell within that level. This replaces $\mathcal{S}_i$ in the rotation above with the head-specific coordinate $\mathcal{S}_i^{(l_h)}$, so the head's queries and keys are rotated using a single 3D surface coordinate at one granularity.

\end{minipage}
\hfill
\begin{minipage}{0.48\textwidth}
    \includegraphics[width=\linewidth]{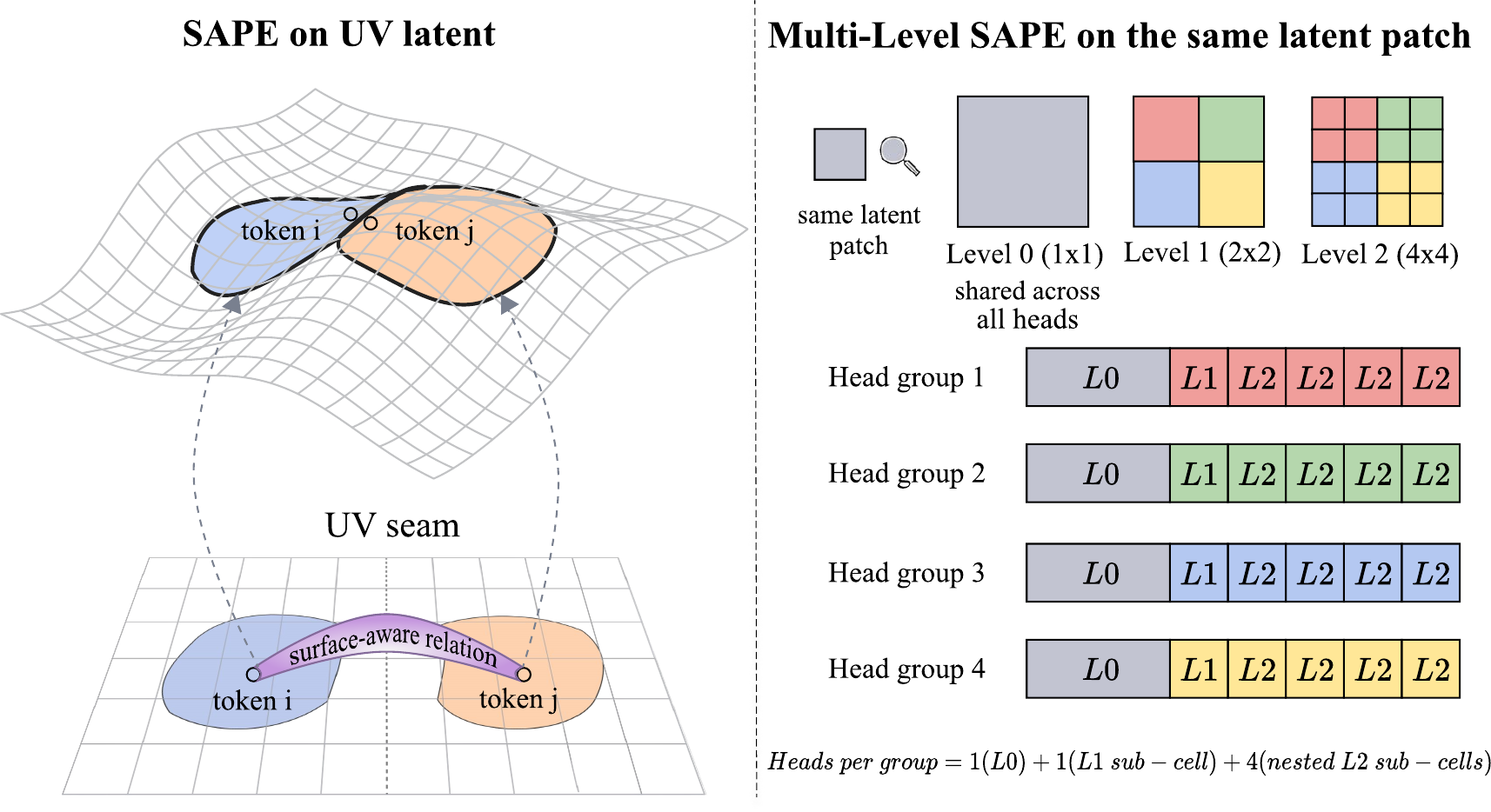}
    \captionof{figure}{Mechanism of SAPE and Multi-Level SAPE. \textbf{Left.} Tokens $i$ and $j$ are adjacent on the mesh but separated onto different UV islands by a seam. SAPE measures their attention distance on the surface, allowing direct interaction across the seam. \textbf{Right.} Multi-Level SAPE reads the CCM at $L$ resolutions and assigns the resulting positional encodings to the $H$ heads, organized into $G$ groups (one cell per head, color denotes group).}
    \label{fig:sape_mechanism}
\end{minipage}

The $H$ heads are organized into $G$ groups, each tied to one sub-cell of the level-$1$ layout. Within a group (for $L=3$), one head reads the level-$0$ coordinate, one head reads the group's level-$1$ sub-cell, and the remaining four heads read the level-$2$ sub-cells nested inside it. The $G$ groups together partition the level-$1$ layout, so every sub-cell at any level is read by exactly one head across the head dimension. The specific values of $L$, $H$, $G$, and the per-level head budget are given in Section~\ref{sec:impl}.

A token's feature vector therefore simultaneously carries surface information at all $L$ granularities, but on disjoint head slices: each head receives a single CCM coordinate rather than a stacked superposition of multiple levels. Since attention is computed per head and then mixed by the output projection, multi-level surface evidence is integrated downstream of attention, while inside attention each head reasons under a clean single-granularity positional metric. No additional latent maps, parameters, or auxiliary geometry branches are introduced. Figure~\ref{fig:sape_mechanism} illustrates how SAPE establishes surface-aware interaction directly on the UV latent map and how each head within a group receives one specific CCM sub-cell.

\subsection{Training and Inference}\label{sec:train}
\noindent\textbf{Training objective.}
We train $v_\theta$ under the rectified flow-matching formulation~\cite{liu2022flow}. Given a clean UV latent $\mathbf{x}$, a noise sample $\boldsymbol{\epsilon}\sim\mathcal{N}(\mathbf{0},\mathbf{I})$, and a timestep $t\in[0,1]$, we form the noisy latent $\mathbf{z}_t = (1-t)\,\mathbf{x} + t\,\boldsymbol{\epsilon}$ and train the network to predict the corresponding flow direction $\boldsymbol{\epsilon}-\mathbf{x}$ via
\begin{equation}
\mathcal{L}_\theta = \mathbb{E}_{t,\mathbf{x},\boldsymbol{\epsilon},\mathbf{c}}\bigl[\,w(t)\,\|v_\theta(\mathbf{z}_t, t, \mathbf{c}) - (\boldsymbol{\epsilon}-\mathbf{x})\|_2^2\,\bigr],
\end{equation}
where $\mathbf{c}$ is the conditioning bundle defined in Section~\ref{sec:cond}. We sample $t$ uniformly from $[0,1]$ and use a uniform reweighting $w(t)\equiv 1$. Note that the CCM is not part of $\mathbf{c}$ and enters the model exclusively through SAPE.

\noindent\textbf{Inference.}
During inference, given a reference image and a 3D mesh, the coarse UV map is produced from the reference image by an off-the-shelf multi-view generator followed by view-to-UV projection, and the CCM is rasterized once from the mesh. Starting from $\mathbf{z}_1\sim\mathcal{N}(\mathbf{0},\mathbf{I})$, we iterate the Euler update for $N$ denoising steps. The conditioning bundle $\mathbf{c}$ and the CCM are deterministic across steps and reused without re-estimation, the final latent is decoded by the frozen Flux VAE to recover the UV texture.


\vspace{-1.0em}
\section{Experiments}\label{sec:experiments}

\subsection{Dataset}

We construct our training set from three large-scale 3D asset datasets:
TexVerse~\cite{zhang2025texverse}, PartNext~\cite{wang2025partnext}, and
Objaverse~\cite{deitke2023objaverse}. After merging and filtering out meshes
with excessive face counts, severely fragmented UV layouts, low-quality UV
parameterizations, or invalid UV-space supervision, we retain approximately
100K textured meshes.

For each mesh we generate a UV mask, a ground-truth UV texture, and
a CCM, all at \(1024 \times 1024\) resolution. We also render four canonical
RGB views at azimuth angles \(\{0^\circ, 90^\circ, 180^\circ, 270^\circ\}\)
with elevation \(5^\circ\). The front view serves as the reference image fed
to DINOv2 and CLIP.

For latent normalization, we precompute the per-position per-channel mean
and standard deviation of VAE-encoded UV latents over the training set, and
apply them during training and inference.

\subsection{Implementation Details}
\label{sec:impl}
Our UV diffusion model is a Flux-style transformer with 6 MMDiT blocks and 12 single-stream blocks, using $H=24$ attention heads with head dimension $d_\text{head}=120$ and axes dimensions $(40, 40, 40)$. For Multi-Level SAPE, we use $L=3$ subdivision levels and partition the $H$ heads into $G=4$ groups of six. Within each group, one head reads the level-0 coordinate, one reads its level-1 sub-cell, and the remaining four read the four level-2 sub-cells nested inside, giving a 4:4:16 ratio across levels. Image conditioning uses DINOv2 ViT-L/14 with registers and CLIP ViT-B/32. For UV encoding and decoding, we reuse the frozen VAE from FLUX.1-dev~\cite{flux2024}, which has $8\times$ spatial downsampling and 64 latent channels. The noisy target UV latent and the VAE-encoded coarse UV are concatenated along the channel dimension, yielding 128 input channels and 64 output channels at the patchify layer.

The coarse UV map is constructed online by projecting rendered views into UV space using the known mesh geometry. At each step, we randomly select $0$, $1$, $2$, or $4$ views, where the zero-view case corresponds to a blank UV map. To simulate realistic view-level inconsistencies, some views are rendered with lighting while others use albedo colors, introducing illumination variation across the projected UV and making it an imperfect conditioning signal rather than a target to be directly copied.

We train on 8 NVIDIA A800 GPUs (Accelerate) with batch size 16 per GPU for 170K steps using 8-bit AdamW ($10^{-5}$ lr, $10^{-4}$ wd, betas $(0.9,0.999)$, clip 1.0), bf16 and gradient checkpointing. 
At inference, we run 28 denoising steps with \texttt{FlowMatchEulerDiscreteScheduler}; coarse UV maps are obtained via Kiss3dGen~\cite{lin2025kiss3dgen} followed by projection to the mesh.

\begin{figure}[t]
\centering
\includegraphics[width=\linewidth]{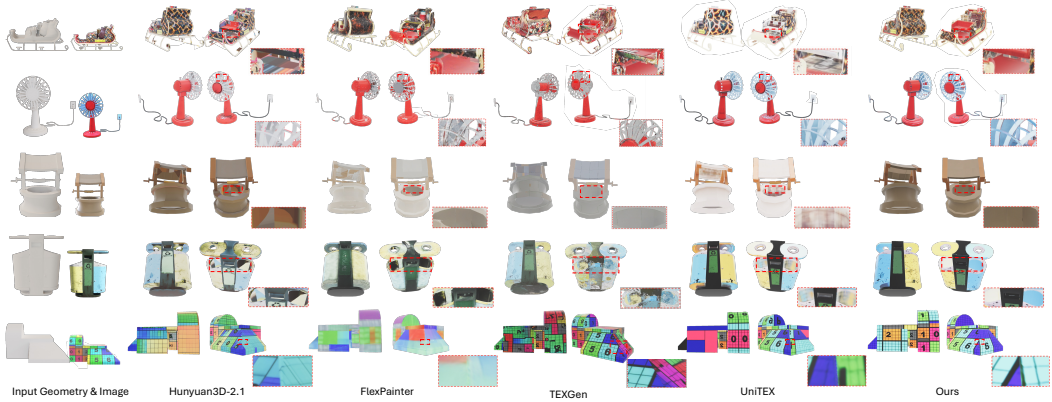}
\caption{Qualitative comparison with state-of-the-art texture generation methods. \textbf{(i) Finer details:} our results preserve sharper high-frequency patterns, e.g., the line strokes in the number block. \textbf{(ii) Occluded regions:} for heavily occluded interiors such as the well and the trash bin, our method produces coherent and plausible content, while others yield blurry or incomplete textures. \textbf{(iii) Robustness to multi-view inconsistency:} by denoising directly in latent UV space, our method avoids being misled by conflicting views. In the fan case, UniTEX incorrectly transfers the back-cover texture onto the front blades, whereas ours preserves the correct appearance.}
\vspace{-0.6em}
\label{fig:comparsion}
\end{figure}

\subsection{Evaluation}

We evaluate on 200 objects randomly sampled from the Google Scanned Objects
(GSO)~\cite{downs2022google}
dataset, which is disjoint from our training data and thus serves as an
out-of-distribution test set. For each object, we render a single
front-facing view as the reference image input for all methods.

For quantitative evaluation, each textured mesh is rendered under albedo
shading from a set of viewpoints providing full directional coverage,
including top-down and bottom-up views. Renderings are compared against
ground-truth albedo images, and we report PSNR and SSIM for per-pixel
fidelity, along with FID~\cite{heusel2017gans} and
KID~\cite{binkowski2018demystifying} for distributional similarity.

\subsection{Comparison with State-of-the-Art Methods}
We compare our method with representative 3D texture generation
baselines. TEXGen~\cite{yu2024texgen} generates mesh textures via a hybrid diffusion
model that couples 2D UV-space convolutions with 3D point cloud attention,
exchanging features between the two representations to achieve surface-aware
texture synthesis. FlexPainter~\cite{yan2024flexipainter} unifies multiple control
signals within a texture generation diffusion framework, supporting
text, image, and multi-modal inputs.
Hunyuan3D-2.1~\cite{hunyuan3d2025hunyuan3d} synthesizes multi-view
images with 3D-aware RoPE in its Paint module and bakes the result into
a UV texture. UniTEX~\cite{liang2025UnitTEX} retains the multi-view
projection stage and completes the resulting partial textured mesh
through a Large Texturing Model that produces a triplane feature volume.
All methods receive the same input meshes and reference images.

\noindent\textbf{Quantitative Results.}
Table~\ref{tab:gso_quantitative} reports the quantitative comparison. Our method achieves the best PSNR and SSIM among all compared
methods, demonstrating superior reference fidelity and reconstruction accuracy.
In addition, our method obtains competitive FID and KID scores, indicating that
it preserves perceptual realism while maintaining stable texture consistency.
Across all metrics, our method achieves the best overall balance between reconstruction fidelity and perceptual quality.

\begin{table}[t]
\centering
\begin{tabular}{lcccc}
\toprule
Method & PSNR $\uparrow$ & SSIM $\uparrow$ & FID $\downarrow$ & KID $\downarrow$ \\
\midrule
TEXGen~\cite{yu2024texgen} & 24.45 & 0.9417 & 44.09 & 34.32 \\
FlexPainter~\cite{yan2024flexipainter} & 23.21 & 0.9463 & 47.04 & 39.25 \\
Hunyuan3D-2.1~\cite{hunyuan3d2025hunyuan3d} & \underline{24.74} & 0.9450 & \underline{39.872} & \underline{33.34} \\
UniTEX~\cite{liang2025UnitTEX} & 24.23 & \underline{0.9504} & 42.11 & 39.21 \\
Ours & \textbf{25.13} & \textbf{0.9527} & \textbf{39.17} & \textbf{30.03} \\
\bottomrule
\end{tabular}
\caption{Quantitative comparison on the GSO dataset. Best results are shown in bold, and second-best results are underlined.}
\vspace{-2.3em}
\label{tab:gso_quantitative}
\end{table}

\noindent\textbf{Qualitative Results.}
All qualitative renderings are produced under identical lighting conditions applied uniformly to all methods.
We compare our method with the strongest baselines on a diverse set of meshes, as illustrated in Figure~\ref{fig:comparsion}. Our results consistently exhibit finer texture details, as reflected in the sharp recovery of high-frequency patterns on the number block, where competing methods tend to blur or misalign thin structures.
In heavily occluded regions, such as the inner walls of the well and the trash bin, our method produces coherent and semantically plausible content, while projection-based pipelines often yield blurry or fragmented artifacts due to limited visible evidence.
The fan example further demonstrates robustness to multi-view inconsistency. UniTEX treats projected multi-view results as the target to be completed via inpainting, and thus transfers the back-cover texture onto the front blades when the views are inconsistent. In contrast, our method operates on the UV latent with conditioning signals, resolving such conflicts through global reasoning and preserving the correct appearance.

\subsection{Ablation Study}
\begin{figure}[t]
\centering
\includegraphics[width=0.7\linewidth]{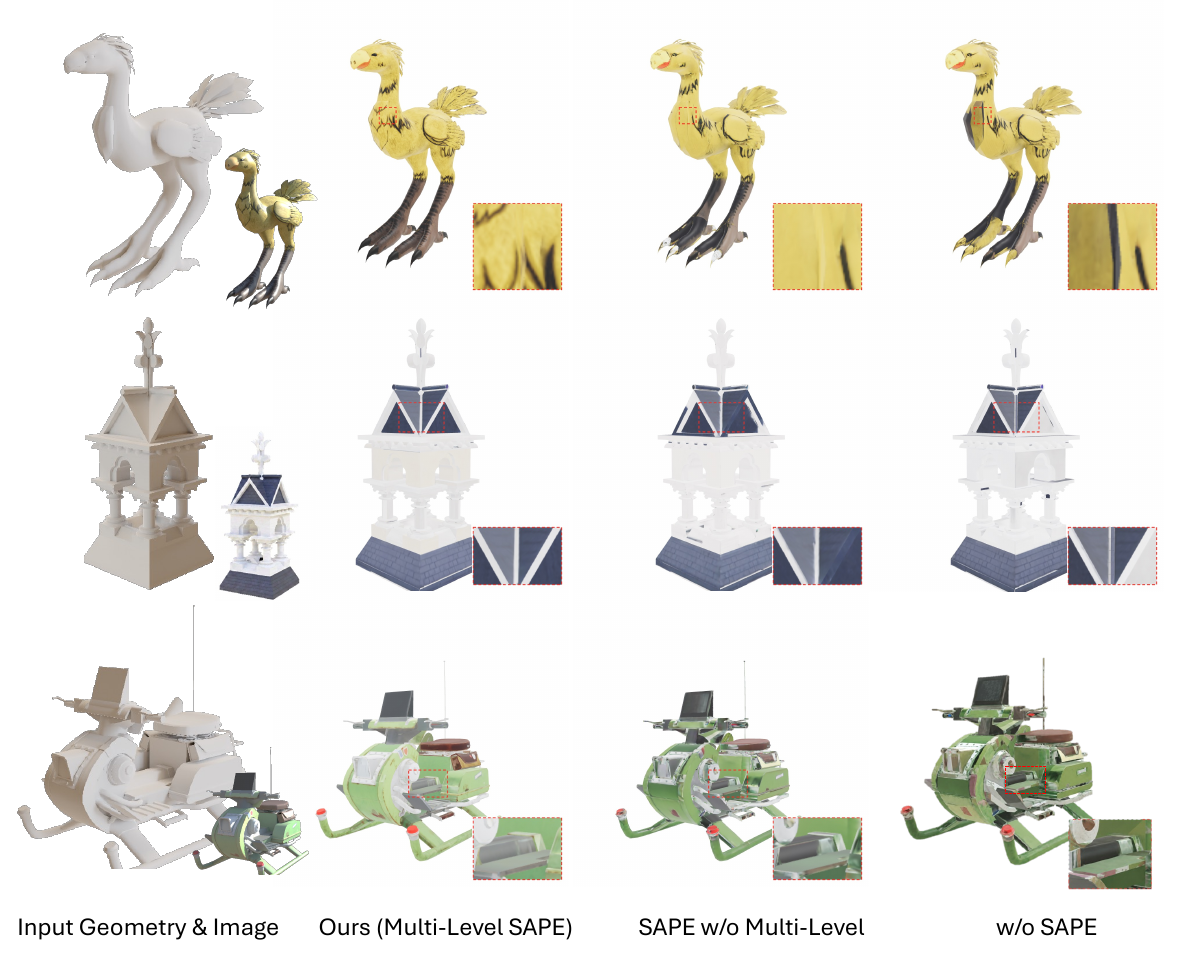}
\caption{Qualitative ablation of Surface-Aware Positional Encoding. \textbf{Without SAPE}, texels that are adjacent on the 3D surface but split into different UV islands cannot interact properly, leading to \textbf{severe appearance divergence} in some regions, as seen in the chocobo case where the black texture from the legs is incorrectly generated on the neck. \textbf{Removing only the multi-level design} preserves cross-island coherence but loses spatial precision, yielding \textbf{blurry textures}.}

\vspace{-1.0em}
\label{fig:ablation_qualitative}
\end{figure}

\begin{minipage}{0.48\textwidth}
\noindent\textbf{Effect of Surface-Aware Positional Encoding.}
To isolate the contribution of SAPE, we replace it with the standard 2D
UV-grid RoPE used by Flux while keeping the rest of the model unchanged,
yielding a pure latent UV-space DiT with no surface coordinate
information in its positional encoding. As shown in
Table~\ref{tab:ablation} and Figure~\ref{fig:ablation_qualitative}, this
\end{minipage}
\hfill
\begin{minipage}{0.5\textwidth}
    \centering
    \resizebox{\columnwidth}{!}{
    \begin{tabular}{lcccc}
    \toprule
    Variant & PSNR $\uparrow$ & SSIM $\uparrow$ & FID $\downarrow$ & KID $\downarrow$ \\
    \midrule
    w/o SAPE                & 23.82 & 0.9376 & 58.42 & 72.36 \\
    SAPE w/o Multi-Level    & \underline{24.36} & \underline{0.9462} & \underline{40.86} & \underline{36.74} \\
    Ours (Multi-Level SAPE) & \textbf{25.13} & \textbf{0.9527} & \textbf{39.17} & \textbf{30.03} \\
    \bottomrule
    \end{tabular}
    }
    \captionof{table}{Ablation of SAPE and its multi-level design on the GSO dataset.}
    \label{tab:ablation}
\end{minipage}

variant drops across all metrics, with the most visible degradation on
regions that surface-disconnected islands. Attention can
only relate tokens by UV-plane proximity, so mesh-adjacent texels split
across charts cannot interact directly, producing seam mismatches and
inconsistent appearance between disconnected UV islands of the same
surface region.

\noindent\textbf{Effect of Multi-Level Design.}
Holding SAPE in place, we ablate only its multi-level structure by tying
all $H$ heads to a single level-0 coordinate
$\mathcal{S}_i = \mathcal{S}_i^{(0)}$. As illustrated in
Table~\ref{tab:ablation} and Figure~\ref{fig:ablation_qualitative}, this
single-level variant retains seam coherence but produces blurry textures.
With every head reading the same coordinate per token, attention cannot
distinguish sub-token positions inside a latent patch, so within-patch
detail collapses to a near-uniform appearance after decoding. Multi-level
coordinates expose finer sub-patch granularities to different heads,
breaking this ambiguity and restoring high-frequency detail.

\section{Conclusion}
We presented DirectUV, an image-conditioned UV texture diffusion framework that generates textures directly in latent UV space, using projected multi-view results only as conditioning signals. The key idea is Surface-Aware Positional Encoding (SAPE), which embeds surface geometry into positional encodings so that attention operates on true surface proximity, enabling coherent synthesis across seams without auxiliary 3D modules. A multi-level design further allows reasoning at multiple spatial scales. DirectUV outperforms other baselines, especially in occluded and view-unseen regions. A remaining limitation is the reliance on UV parameterization quality, which future work may address.


\bibliographystyle{IEEEtran}
\bibliography{references.bib}

\newpage
\appendix
\section*{Appendix}
\section{Coarse UV as Soft Conditioning}
\label{sec:appendix_soft_prior}

The coarse UV map assembled from multi-view projections provides a surface-aligned layout cue, but it can carry inaccurate or misleading color in regions where the projected views do not faithfully represent the true surface appearance. DirectUV treats this map as a soft conditioning signal and does not directly copy its values into the final texture; instead, the denoising process can deviate from the coarse UV when the model's learned prior suggests a more coherent appearance.

\begin{figure}[h]
\centering
\includegraphics[width=\linewidth]{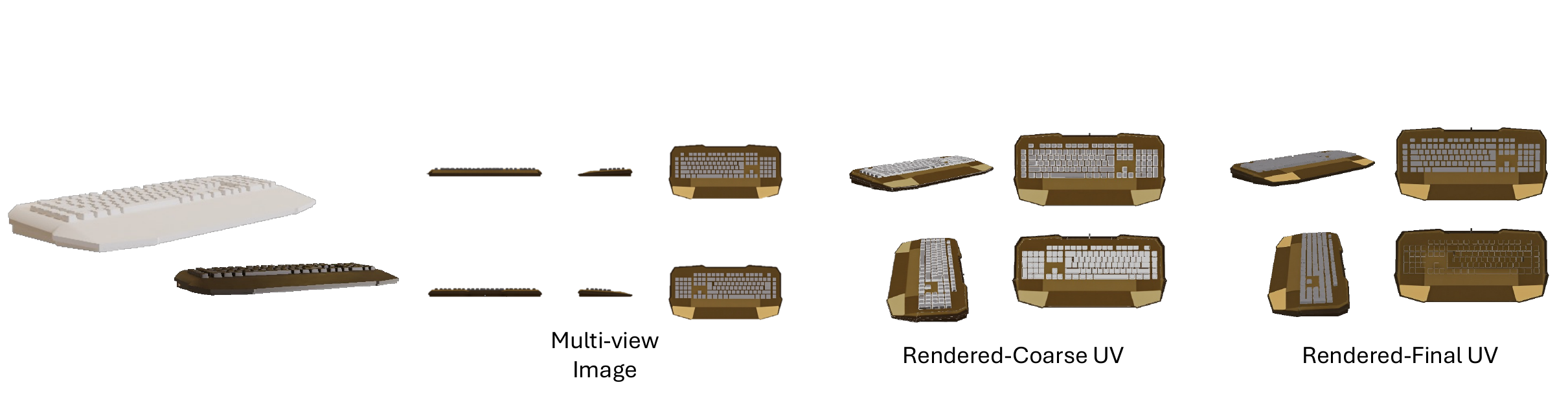}
\caption{The keyboard underside appears white in the coarse UV (center) despite being covered by the bottom-view projection, but is correctly recovered as dark metallic casing in the final texture (right), consistent with the reference image. DirectUV's learned prior overrides the inaccurate coarse UV values rather than copying them.}

\label{fig:no_trust_coarse}
\end{figure}

Figure~\ref{fig:no_trust_coarse} illustrates this on a keyboard example. The coarse UV renders the underside of the chassis as white, yet the reference image and surrounding surface clearly indicate a continuous dark metallic casing. The final texture correctly assigns the metal material to the bottom face, overriding the white coarse UV through global reasoning over the full surface. This behavior confirms that the coarse UV guides rather than constrains the generation.

We quantify this behavior in Table~\ref{tab:coarse_correction}. We use the same
coarse UV map directly as the mesh texture, render it from the evaluation
cameras, and compare it against DirectUV \emph{only} at pixels corresponding to
projection-covered texels. Restricting the comparison in this way excludes any
gain that would come from merely filling empty texels, so the remaining
improvement isolates how DirectUV treats projected content that already exists.
DirectUV improves all three metrics, showing that it corrects the projected
values rather than copying them.

\begin{table}[h]
\centering
\caption{DirectUV improvement over rendering the coarse UV map directly as the
mesh texture, evaluated only at coarse-covered pixels.}
\begin{tabular}{lccc}
\toprule
Region & $\Delta$PSNR $\uparrow$ & $\Delta$SSIM $\uparrow$ & LPIPS reduction $\uparrow$ \\
\midrule
Coarse-covered & +1.11 & +0.0301 & +0.03 \\
\bottomrule
\end{tabular}
\label{tab:coarse_correction}
\end{table}

\section{Robustness to Coarse UV Completeness}
\label{sec:appendix_view_count}

\begin{figure}[h]
\centering
\includegraphics[width=\linewidth]{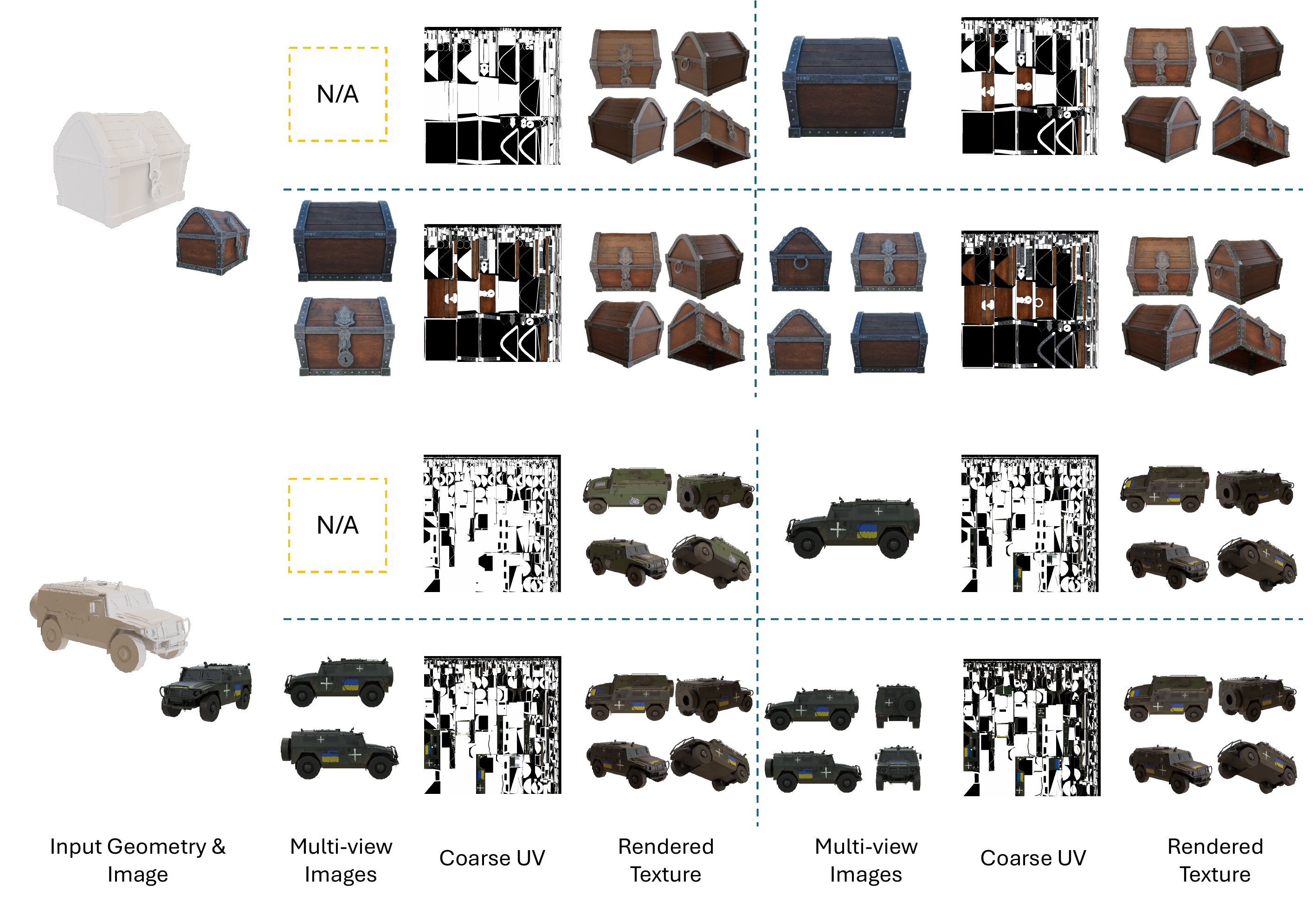}
\caption{Texture generation results under varying coarse UV completeness
at inference time ($0$, $1$, $2$, and $4$ projected views). The reference
image and mesh are identical across all cases.}
\label{fig:view_count_qualitative}
\end{figure}

During training, the number of views used to construct the coarse UV
is randomly sampled from $\{0, 1, 2, 4\}$ at each step, exposing the
model to a wide range of coarse UV completeness and preventing it from
over-relying on any single input quality level. Figure~\ref{fig:view_count_qualitative}
shows that this strategy yields a model robust to incomplete coarse UV
inputs at inference time. Even in the zero-view setting where the coarse
UV is entirely blank, DirectUV produces plausible textures by relying on
the reference image and geometry-conditioned denoising alone, confirming
that the coarse UV is treated as a soft conditioning signal rather than
a hard initialization.

Table~\ref{tab:view_count} reports the corresponding quantitative behavior,
varying only the number of projected views while keeping all other inputs
fixed. The first projected view provides most of the PSNR/SSIM improvement,
after which these metrics largely saturate, while FID and KID continue to
improve with additional views. This indicates that a single view supplies most
of the pixel-level alignment, whereas further views mainly enrich
reference-specific detail and distributional quality.

\begin{table}[h]
\centering
\caption{Effect of the number of projected views used to construct the coarse UV.}
\begin{tabular}{lcccc}
\toprule
Projected views & PSNR $\uparrow$ & SSIM $\uparrow$ & FID $\downarrow$ & KID $\downarrow$ \\
\midrule
0 & 23.07 & 0.9116 & 70.92 & 86.97 \\
1 & 25.08 & 0.9521 & 50.43 & 57.09 \\
2 & 25.11 & 0.9523 & 42.17 & 40.28 \\
4 & \textbf{25.13} & \textbf{0.9527} & \textbf{39.17} & \textbf{30.03} \\
\bottomrule
\end{tabular}
\label{tab:view_count}
\end{table}

\section{Positional Injection versus Auxiliary Geometry Conditioning}
\label{sec:appendix_geometry_injection}

SAPE uses surface coordinates in the positional mechanism rather than in a
conditioning branch. To determine whether its gains arise from this placement
or simply from access to the CCM, we train an identical-backbone variant that
concatenates the normalized CCM with the latent spatial channels while
retaining standard 2D UV RoPE. All other components, training data, and
conditioning signals remain fixed.

Table~\ref{tab:geometry_injection} shows that channel concatenation provides
only modest gains. Although the backbone can use the auxiliary geometry, it
must implicitly infer the coordinate relationships among UV tokens. Using
the same coordinates to define positional relations in self-attention yields
substantially larger gains across all metrics at negligible computational cost.
The comparison between Single-Level and Multi-Level SAPE further isolates the
benefit of sub-patch surface sampling.

\begin{table}[h]
\centering
\caption{Controlled comparison of geometry-injection mechanisms under an
identical backbone, training set, and conditioning pipeline. Only the way the
CCM enters the model differs.}
\resizebox{\linewidth}{!}{
\begin{tabular}{llcccc}
\toprule
Variant & Geometry injection & PSNR $\uparrow$ & SSIM $\uparrow$ & FID $\downarrow$ & KID $\downarrow$ \\
\midrule
w/o SAPE (2D RoPE) & None & 23.82 & 0.9376 & 58.42 & 72.36 \\
CCM concatenation + 2D RoPE & Latent-channel concatenation & 23.97 & 0.9395 & 50.62 & 66.68 \\
Single-Level SAPE & CCM coordinates in RoPE & \underline{24.36} & \underline{0.9462} & \underline{40.86} & \underline{36.74} \\
Multi-Level SAPE & Multi-level CCM coordinates in RoPE & \textbf{25.13} & \textbf{0.9527} & \textbf{39.17} & \textbf{30.03} \\
\bottomrule
\end{tabular}
}
\label{tab:geometry_injection}
\end{table}

\section{Seam Consistency}
\label{sec:appendix_seam}

UV seams separate texels that are adjacent on the mesh surface. We therefore
evaluate texture quality specifically near seams. We identify mesh edges whose
incident faces are continuous in 3D but map to different UV islands, then
sample matched strips just inside both islands using identical surface
interpolation parameters. The strips are joined along their virtual surface
adjacency and packed into fixed-format per-object atlases for comparison with
the corresponding ground truth.

Alongside PSNR, SSIM, FID, and KID on these atlases, we report a
\emph{GT-referenced seam discrepancy}. For matched texels one texel inside each
island, we compute the mean absolute cross-seam RGB jump in the prediction and
ground truth, then average the absolute difference between the two jumps over
all seam samples. With RGB normalized to $[0,1]$, the metric also lies in
$[0,1]$, and lower is better. Referencing the ground truth avoids penalizing
genuine texture discontinuities or rewarding over-smoothed predictions that
remove all cross-seam contrast.

Table~\ref{tab:seam} shows that channel concatenation modestly improves over 2D
RoPE: access to 3D coordinates helps, but does not overcome the positional
discontinuity between UV islands. Single-Level SAPE substantially improves
every seam metric by using these coordinates in attention, and Multi-Level
SAPE improves further, indicating that sub-patch surface locations are
particularly useful near seam boundaries.

Because seam neighborhoods are thin, high-frequency strips that directly
expose cross-island discontinuities, their absolute scores are much lower than
global metrics and should be interpreted comparatively. The low 2D RoPE
baseline reflects the failure mode SAPE is designed to address.

\begin{table}[h]
\centering
\caption{Texture quality on matched UV-seam regions. All variants share the
same backbone, training set, and conditioning.}
\resizebox{\linewidth}{!}{
\begin{tabular}{lccccc}
\toprule
Variant & Seam PSNR $\uparrow$ & Seam SSIM $\uparrow$ & Seam FID $\downarrow$ & Seam KID $\downarrow$ & GT-referenced discrepancy $\downarrow$ \\
\midrule
w/o SAPE (2D RoPE) & 11.30 & 0.2448 & 109.92 & 225.11 & 0.18 \\
CCM concatenation + 2D RoPE & 12.11 & 0.3412 & 91.43 & 183.24 & 0.14 \\
Single-Level SAPE & \underline{16.70} & \underline{0.6555} & \underline{60.59} & \underline{71.60} & \underline{0.10} \\
Multi-Level SAPE & \textbf{21.35} & \textbf{0.8790} & \textbf{48.50} & \textbf{43.28} & \textbf{0.09} \\
\bottomrule
\end{tabular}
}
\label{tab:seam}
\end{table}

\section{Visible and View-Unseen Regions}
\label{sec:appendix_visible_unseen}

The coarse UV provides texture evidence only for surfaces observed by the
projection cameras. Using their union coverage mask, we label a surface texel
as \emph{visible} if it receives evidence from at least one projection and as
\emph{view-unseen} otherwise. View-unseen regions include interior cavities and
undersides that are difficult to capture from standard exterior viewpoints.
We report mask-aware PSNR, SSIM, and LPIPS, omitting FID and KID because they
measure image-set distributions rather than errors within a spatial mask.

Table~\ref{tab:visible_unseen} reports gains over the 2D UV RoPE baseline using
identical regional masks. Channel concatenation yields only small improvements
in either regime, whereas both SAPE variants perform substantially better.
Their largest gains occur on view-unseen surfaces, where the model has no
projected texture evidence and must rely on reference features, surface
coordinates, and the learned prior.

These comparisons keep the coarse UV and all other conditioning fixed. Because
SAPE's largest gains occur where the coarse UV provides no evidence, they
cannot be explained by that condition; its separate contribution is measured
in Table~\ref{tab:view_count}.

\begin{table}[h]
\centering
\caption{Visible and view-unseen region breakdown, reported as gains over the
2D UV RoPE baseline under identical regional masks.}
\begin{tabular}{llccc}
\toprule
Variant & Region & $\Delta$PSNR $\uparrow$ & $\Delta$SSIM $\uparrow$ & LPIPS reduction $\uparrow$ \\
\midrule
CCM concatenation + 2D RoPE & Visible & +0.08 & +0.0014 & +0.01 \\
CCM concatenation + 2D RoPE & Unseen & +0.17 & +0.0151 & +0.01 \\
Single-Level SAPE & Visible & +0.14 & +0.0268 & +0.03 \\
Single-Level SAPE & Unseen & +0.85 & +0.0297 & +0.04 \\
Multi-Level SAPE & Visible & \textbf{+1.14} & \textbf{+0.0578} & \textbf{+0.05} \\
Multi-Level SAPE & Unseen & \textbf{+1.39} & \textbf{+0.0948} & \textbf{+0.05} \\
\bottomrule
\end{tabular}
\label{tab:visible_unseen}
\end{table}

\section{Application to Generated Meshes}
\label{sec:appendix_in_the_wild}

\begin{figure}[h]
\centering
\includegraphics[width=\linewidth]{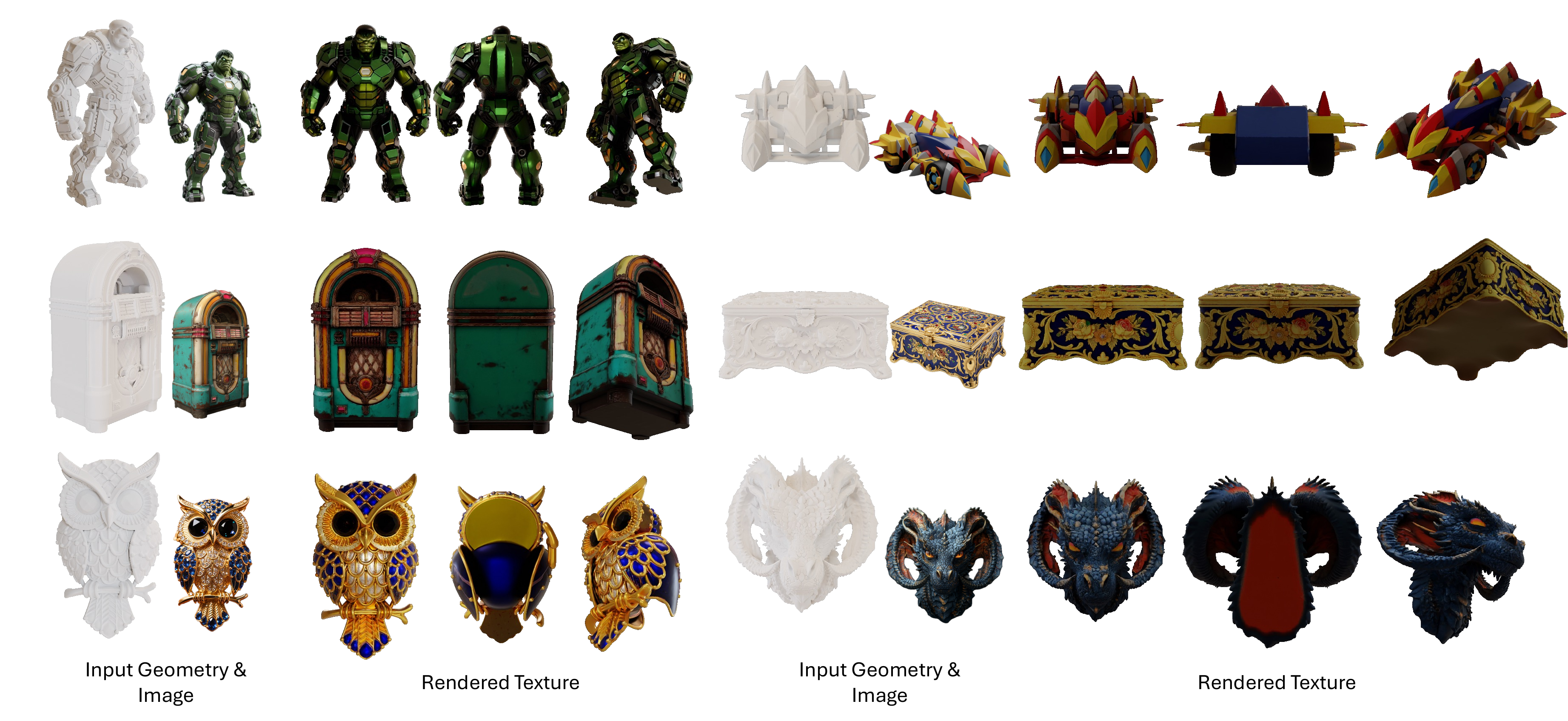}
\caption{Texture generation on meshes produced by image-to-3D shape generators, demonstrating DirectUV in a complete image-to-textured-mesh pipeline. Each block shows the input reference image alongside the generated texture rendered from multiple viewpoints.}
\label{fig:in_the_wild}
\end{figure}

To evaluate DirectUV in the context of current 3D content creation workflows, we apply it to meshes produced by recent image-to-3D shape generators like Tripo and Rodin, forming a complete image-to-textured-mesh pipeline. Given a reference image, the shape generator produces a mesh, and DirectUV textures it conditioned on the same image. Figure~\ref{fig:in_the_wild} shows results across diverse objects, demonstrating that DirectUV handles the varying topologies and UV layouts produced by generative shape models and integrates naturally into an end-to-end production setting. These meshes use the UV parameterizations automatically exported by their platforms, without re-unwrapping or manual cleanup such as island merging, seam repair, or fragment deletion.

\section{Robustness to UV Fragmentation}
\label{sec:appendix_fragmentation}

Island count measures how extensively a surface is partitioned into
disconnected UV charts. Because GSO mainly contains meshes with relatively few
islands, we also evaluate held-out TexVerse objects, extending the tested range
beyond 50{,}000 islands.

Table~\ref{tab:fragmentation} reports quality by island count. For GSO groups
above 10 islands, PSNR remains between 25.73 and 28.27 dB; across held-out
TexVerse groups spanning 10k to more than 50k islands, it remains between 26.00
and 28.57 dB. PSNR, SSIM, and LPIPS show no consistent degradation as island
count increases, supporting robustness to chart-count fragmentation over the
tested range. Section~\ref{sec:appendix_failure} presents a complementary
boundary case with poor local chart geometry, including tiny slivers, corner
patches, and inefficient packing. Thus, DirectUV tolerates high chart counts in
the evaluated layouts, although severely ill-conditioned local
parameterization can still reduce fine-detail quality.

FID and KID are set-level metrics sensitive to subgroup size and composition,
so we do not compare them across groups. The GSO bin above 1{,}000 islands
contains only six objects (48 renders), for which FID worsens while KID
improves. We therefore assess the fragmentation trend using the per-image
metrics.

\begin{table}[h]
\centering
\caption{Texture quality as a function of UV atlas fragmentation, measured by
island count. The held-out TexVerse subsets extend the tested range well beyond
what GSO covers.}
\resizebox{\linewidth}{!}{
\begin{tabular}{llcccccc}
\toprule
Dataset & Island count & Objects & PSNR $\uparrow$ & SSIM $\uparrow$ & LPIPS $\downarrow$ & FID $\downarrow$ & KID $\times 10^4$ $\downarrow$ \\
\midrule
GSO & 1--10 & 77 & 23.19 & 0.9411 & 0.07 & 59.76 & 50.66 \\
 & 11--50 & 48 & 26.31 & 0.9597 & 0.04 & 82.36 & 28.21 \\
 & 51--100 & 23 & 25.73 & 0.9562 & 0.04 & 91.75 & 31.97 \\
 & 101--1000 & 46 & 26.47 & 0.9616 & 0.04 & 73.14 & 44.36 \\
 & $>$1000 & 6 & 28.27 & 0.9637 & 0.03 & 153.78 & 8.97 \\
\midrule
TexVerse & 10k--20k & 220 & 27.09 & 0.9675 & 0.04 & 59.46 & 19.25 \\
(held out & 20k--50k & 134 & 26.00 & 0.9673 & 0.04 & 84.08 & 25.43 \\
from training) & $>$50k & 64 & 28.57 & 0.9706 & 0.04 & 102.28 & 55.69 \\
\bottomrule
\end{tabular}
}
\label{tab:fragmentation}
\end{table}

\section{Failure Cases}
\label{sec:appendix_failure}
\begin{figure}[h]
\centering
\includegraphics[width=\linewidth]{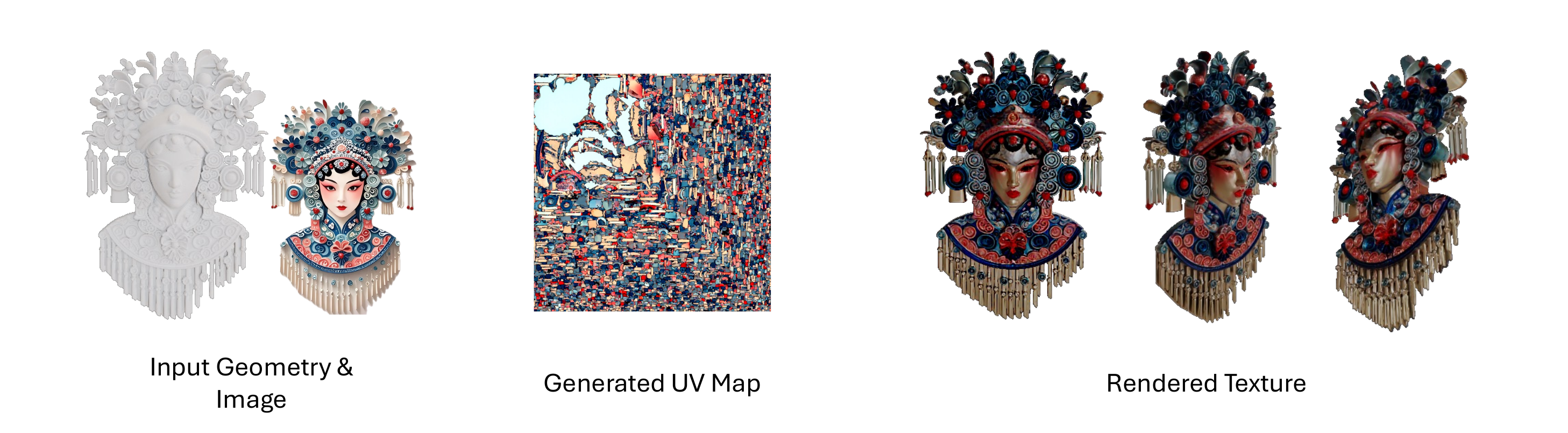}
\caption{Failure cases on heavily fragmented UV layouts. The UV map
(top row) is split into a large number of small islands and slivers
with substantial unused border space, leading to blurred or
color-leaked regions in the wrapped texture (bottom row).}
\label{fig:failure_fragmented}
\end{figure}

DirectUV is sensitive to the quality of the underlying UV parameterization. On meshes with highly fragmented UV layouts—composed of many small slivers and corner patches—the per-token surface coordinate $\mathcal{S}_i$ aggregates information over relatively large surface regions for each fragment, resulting in a coarse CCM signal. In such cases, the multi-level subdivision is unable to fully recover fine-grained local detail. Figure~\ref{fig:failure_fragmented} shows representative examples where the rendered textures exhibit blurriness and color leakage in these regions. 

These failure modes arise from the UV parameterization and mesh quality, rather than from the SAPE design itself. We view this limitation as an opportunity for future work, such as UV-aware preprocessing or jointly learned tokenization strategies tailored to UV-space data.

\section{Inference Runtime and Memory}
\label{sec:appendix_runtime}

We measure 28-step denoising at batch size 1 on a single A100-80GB GPU, using
the same backbone across variants to isolate positional-encoding cost.
Table~\ref{tab:runtime} reports latency and peak GPU memory. Single-Level SAPE
has no measurable overhead over standard 2D UV RoPE, while Multi-Level SAPE
adds only a small latency increase and no additional memory cost. SAPE has no
learnable parameters and only changes the coordinates passed to the existing
rotary embedding.

Texture resolution determines the number of UV tokens and dominates the cost,
whereas mesh resolution mainly affects the one-time rasterization used to
construct the CCM and coarse UV. The reported numbers cover denoising only; the
multi-view generation stage is an off-the-shelf component shared with
projection-based pipelines.

\begin{table}[h]
\centering
\caption{Denoising time and peak GPU memory under an identical backbone. Each
cell reports latency / peak memory.}
\begin{tabular}{lccc}
\toprule
Resolution & 2D UV RoPE & Single-Level SAPE & Multi-Level SAPE \\
\midrule
$512^2$ & 13.06 s / 15.20 GiB & 13.05 s / 15.20 GiB & 13.38 s / 15.20 GiB \\
$1024^2$ & 35.99 s / 18.04 GiB & 35.98 s / 18.04 GiB & 36.40 s / 18.04 GiB \\
\bottomrule
\end{tabular}
\label{tab:runtime}
\end{table}

\section{More Results}
\label{sec:appendix_more_results}

\begin{figure}[!h]
\centering
\includegraphics[width=0.99\linewidth]{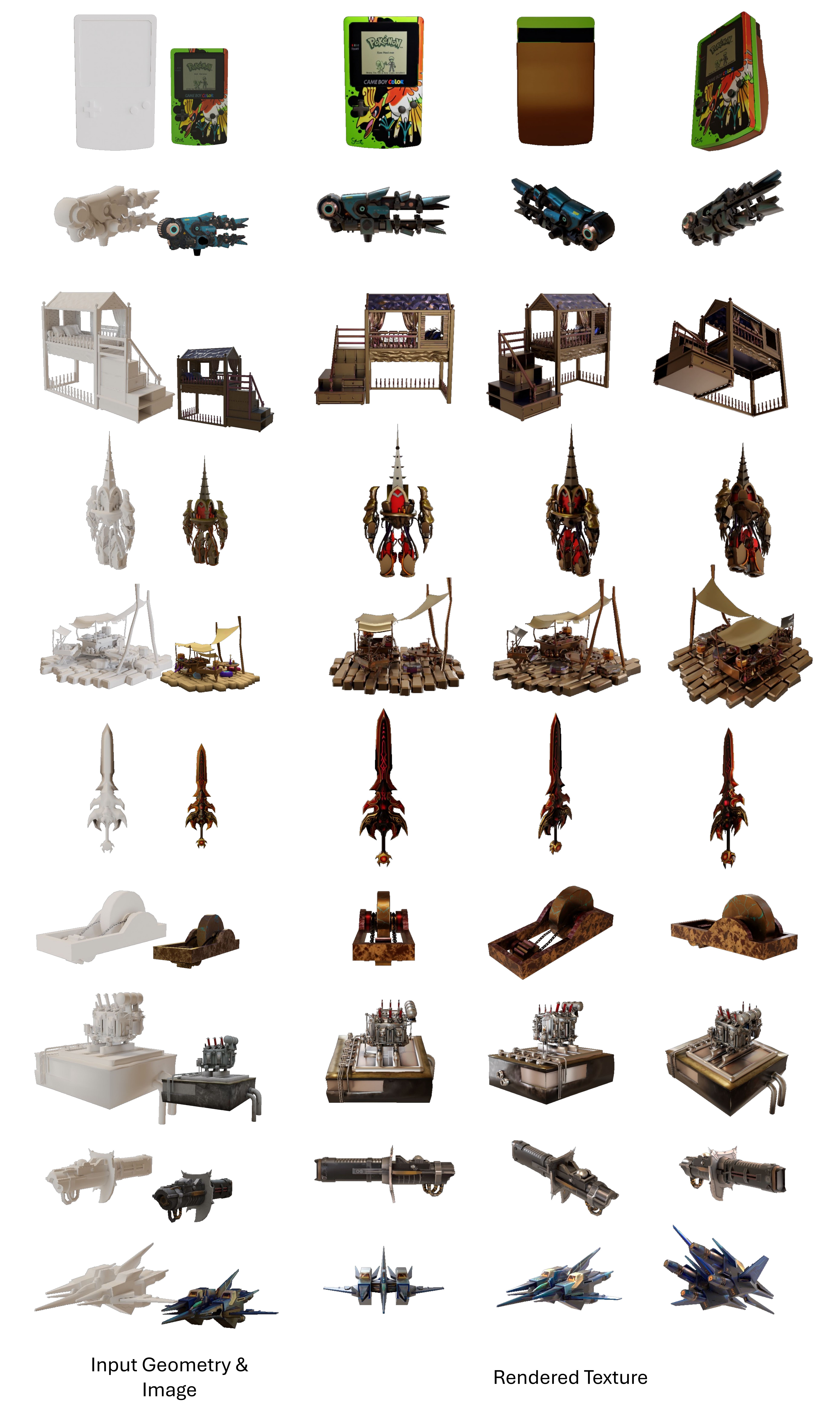}
\vspace{-1.3em}
\caption{Additional texture generation results.}
\label{fig:more_results}
\end{figure}

We present additional texture generation results in
Figure~\ref{fig:more_results}, covering a diverse range of object
categories and mesh complexities.

\end{document}